\documentclass[conference]{IEEEtran}
\IEEEoverridecommandlockouts

\usepackage{cite}
\usepackage{amsmath,amssymb,amsfonts}
\usepackage{algorithm}
\usepackage{algorithmic}
\usepackage[dvipdfmx]{graphicx}
\usepackage{textcomp}
\usepackage{xcolor}
\usepackage{comment}
\usepackage{bm}
\def\BibTeX{{\rm B\kern-.05em{\sc i\kern-.025em b}\kern-.08em
    T\kern-.1667em\lower.7ex\hbox{E}\kern-.125emX}}

\begin{document}

\title{A Tunable Model for Graph Generation\\Using LSTM and Conditional VAE}

\author{\IEEEauthorblockN{1\textsuperscript{st} Shohei Nakazawa}
\IEEEauthorblockA{\textit{Graduate School of Engineering} \\
\textit{Nagaoka University of Technology}\\
Nagaoka, Niigata, Japan\\
s173160@stn.nagaokaut.ac.jp}
\and
\IEEEauthorblockN{2\textsuperscript{nd} Yoshiki Sato}
\IEEEauthorblockA{\textit{Graduate School of Engineering} \\
\textit{Nagaoka University of Technology}\\
Nagaoka, Niigata, Japan\\
s171039@stn.nagaokaut.ac.jp}
\and
\IEEEauthorblockN{3\textsuperscript{rd} Kenji Nakagawa}
\IEEEauthorblockA{\textit{Graduate School of Engineering} \\
\textit{Nagaoka University of Technology}\\
Nagaoka, Niigata, Japan\\
nakagawa@nagaokaut.ac.jp}
\and
\IEEEauthorblockN{4\textsuperscript{th} Sho Tsugawa}
\IEEEauthorblockA{\textit{Faculty of Engineering, Information and Systems} \\
\textit{University of Tsukuba}\\
Tsukuba, Ibaraki, Japan\\
s-tugawa@cs.tsukuba.ac.jp}
\and
\IEEEauthorblockN{5\textsuperscript{th} Kohei Watabe}
\IEEEauthorblockA{\textit{Graduate School of Engineering} \\
\textit{Nagaoka University of Technology}\\
Nagaoka, Niigata, Japan\\
k\_watabe@vos.nagaokaut.ac.jp}
}

\maketitle

\begin{abstract}
With the development of graph applications, generative models for graphs have been more crucial.
Classically, stochastic models that generate graphs with a pre-defined probability of edges and nodes have been studied.
Recently, some models that reproduce the structural features of graphs by learning from actual graph data using machine learning have been studied.
However, in these conventional studies based on machine learning, structural features of graphs can be learned from data, but it is not possible to tune features and generate graphs with specific features.
In this paper, we propose a generative model that can tune specific features, while learning structural features of a graph from data.
With a dataset of graphs with various features generated by a stochastic model, we confirm that our model can generate a graph with specific features. 
\end{abstract}

\begin{IEEEkeywords}
Graph generation, Conditional VAE, LSTM
\end{IEEEkeywords}

\section{Introduction}\label{introduction}

Graph generation has a wide range of applications, such as communication networks, transportation systems, databases, molecules chemistry, etc.  
The generative models for graphs can be categorized into two types: statistical models and machine-learning-based models.
The traditional statistical models are possible to generate a graph that reproduces a single-aspect feature (e.g. scale-free) focused by the designer of the model. 
The generative models for graphs using machine learning technology learn features from graph data and try to reproduce features according to the data in every single aspect. 
However, to our best knowledge, the existing models based on machine learning cannot tune a specific feature while learning multi-aspect features from graph data. 

In this paper, we propose a generative model for graphs that enables to tune specific structural features using Depth First Search (DFS) code and Conditional Variational Auto Encoder (CVAE). 
Unlike the traditional statistical models, the proposed model learns multi-aspect features from graph data.
Though the conventional machine-learning-based models try to reproduce similar structural features of the given graphs, the proposed model can generate a graph with specified features.

\section{Problem Formulation}\label{problem_formulation}

\begin{figure}[tb]
	\begin{center}
		\includegraphics[width=\linewidth]{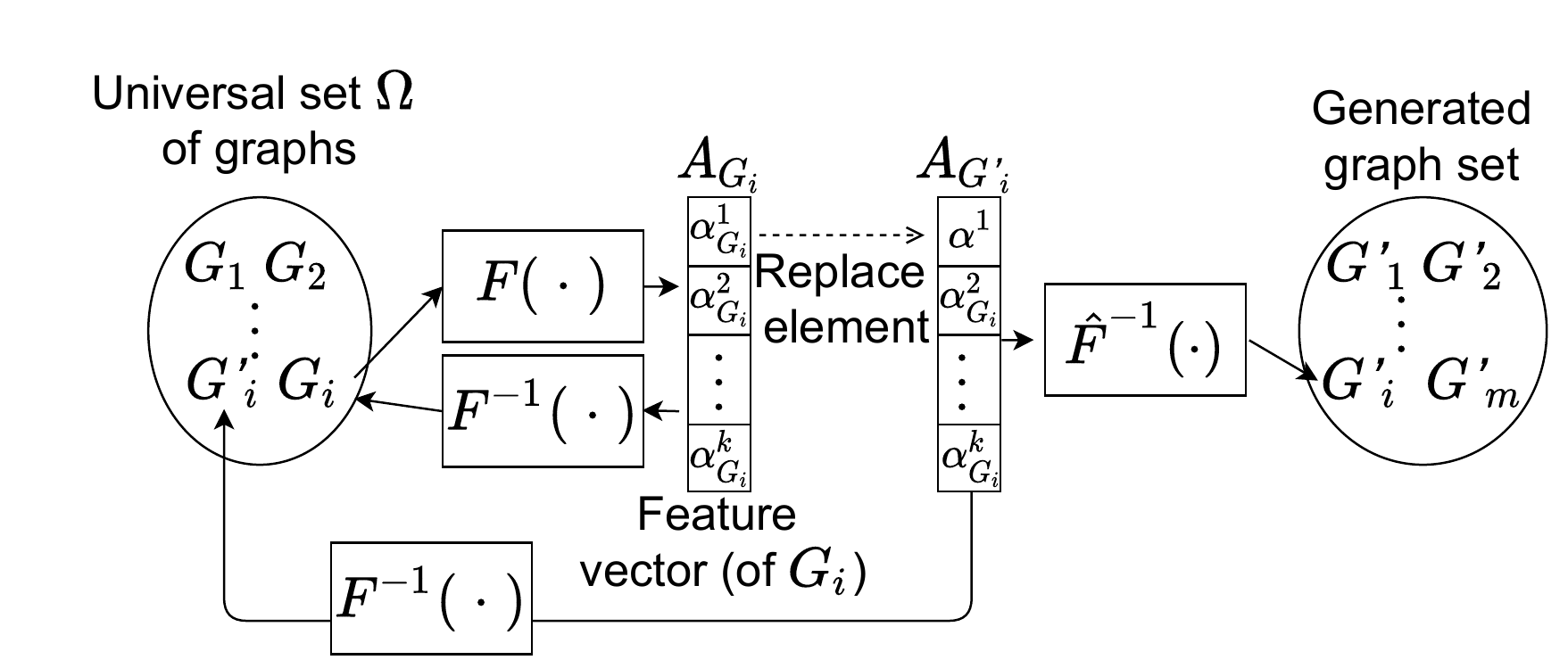}
	\end{center}
	\vspace*{-3ex}
	\caption{Problem formulation}
	\label{problem_formulation_pic}
	\vspace*{-3ex}
\end{figure}

In this paper, we formulate the problem to generate a graph with specified features as the estimation of the mapping from graphs to feature vectors of graphs (see Fig.\ref{problem_formulation_pic}).
Let us denote the universal set of all graphs by $\bm{\Omega}=\{G_1, G_2, \dots \}$. 
We denote a mapping by $F(\cdot)$, and a graph $G_i\in \Omega$ is mapped to a feature vector $A_{G_i}\in \mathbb{R}^k$ by $F(G_i)=A_{G_i}=[\alpha_{G_i}^1 \alpha_{G_i}^2 \cdots \alpha_{G_i}^k]^T$.
The $j$th element $\alpha_{G_i}^j$ of the vector $A_{G_i}$ expresses a feature of graph $G_i$ (i.e. the number of nodes/edges, scaling exponent of the degree distribution, clustering coefficient, etc.).
We tackle the estimation problem to find $\hat{F}^{-1}(\cdot)$ that approximates $F^{-1}(\cdot)$, using a subset of $\Omega$.
By solving this problem, it is possible to generate $G_i'$ with the feature vector $A_{G_i'}$ in which any element of the vector $A_{G_i}$ is replaced by an arbitrary value.

\section{DFS Code}\label{graphgen}

GraphGen~\cite{Goyal2020} is a kind of conventional generative model for graphs using machine learning, and it learns sequence data to which converted from a graph using DFS code.
DFS code, a kind of encoding method for graphs, encodes a graph into a sequence of edges using a depth-first search on the graph. 
In DFS code, first of all, timestamps are added to all nodes from 0 in order of the depth-first search. 
By adding timestamp for all nodes, edge $e=(u, v)$ can be annotated as 5-tuple $(t_u, t_v, L(u), L(e), L(v))$, where $t_u$ and $L(\cdot)$ denote a timestamp of node $u$ and a label of a edge/node, respectively. 
DFS code encodes the graph so that the sequence composed of consecutive neighboring edges as much as possible while maintaining the order of timestamps of the nodes.
Finally, the graph is represented as sequence of 5-tuple $(t_u, t_v, L(u), L(e), L(v))$ by DFS code.

\section{Proposed Model}\label{proposed}

We propose a generative model for graphs that can tune specific features using DFS code and CVAE. 
The proposed model learns a sequence dataset $\bm{S}=\{S_1, S_2, \dots\}=F_{\rm minDFS}(\bm{G})$ that is encoded from a graph dataset $\bm{G}$ with DFS code. 
An end element of a sequence is a vector that represents EOS. 
Nodes are labeled reciprocal of their degree and edges are not labeled. 
We define functions that calculate a scaling exponent of a degree distribution and a clustering coefficient by $F_{\rm deg}(\cdot)$ and $F_{\rm cluster}(\cdot)$, respectively. 
We construct a conditional vector set ${\bm C}=\{c_1, \dots, c_m\}=\{[F_{\rm deg}(G_1)\; F_{\rm cluster}(G_1)]^T, \dots, [F_{\rm deg}(G_m)\; F_{\rm cluster}(G_m)]^T\}$ by using these functions for the graph dataset.

\begin{figure}[tb]
  \vspace{0ex}
	\begin{center}
		\includegraphics[width=\linewidth]{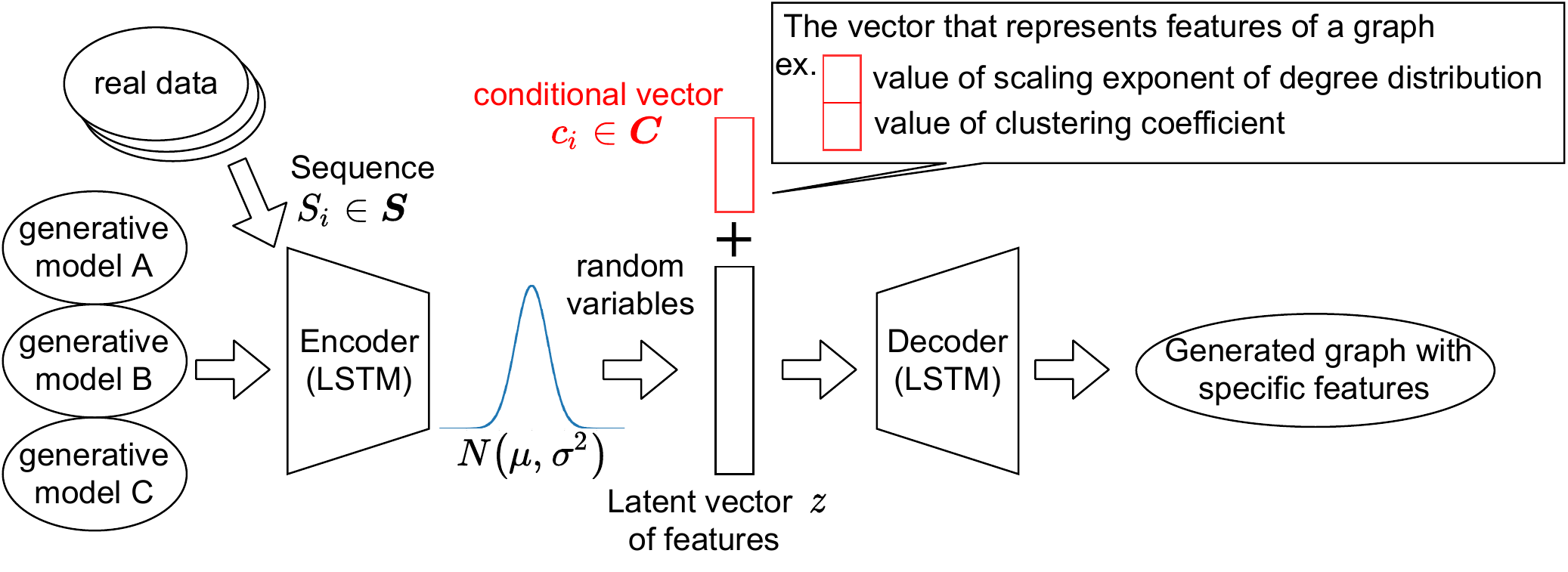}
	\end{center}
  \vspace{-3ex}
	\caption{The proposed model}
	\label{model_pic}
	\vspace{-5mm}
\end{figure}

The proposed model is composed of CVAE with Long Short Term Memory (LSTM) based encoder and decoder (see Fig.~\ref{model_pic}). 
The proposed model is trained with the sequence dataset ${\bm S}$ and conditional vector set $\bm{C}$.

{\bf{Encoder: }} Encoder learns sequences $S_i$ and maps them to a multivariate normal distribution according to features of graphs. 
A latent vector $z$ is sampled from the distribution.
To learn a condition of a graph, we input a condition vector $c_i$ with 5-tuple of the sequence into each LSTM block. 

{\bf{Decoder: }} Decoder learns to map $j$th 5-tuples $s_j \in S_i$ and a conditional vector $c_i$ to next 5-tuples $s_{j+1}$. 
A conditional vector $c_i$ is concatenated with the 5-tuples and input into all LSTM blocks. 
We input a latent vector $z$ into LSTM blocks concatenating with a hidden state.

When we generate a graph using the proposed model, we give a sampled latent vector and a conditional vector whose elements are tuned to specific values to the encoder and the decoder. 
By giving a conditional vector, the decoder recursively generates a sequence of 5-tuples according to the conditional vector.

\section{Experiments}\label{experimentation}
We verified that the proposed model can learn structural features from graph data and generate a graph with specific features.
As an initial study of the tunable model, we utilize a dataset generated by a kind of statistical models called Connecting Nearest Neighbor model~\cite{Vazquez2003a}, for training, though our ultimate goal is to generate a graph with arbitrary features from real graph data. 
The statistical models easily supply massive homogeneous data with well-known features (scale-free, clustering, etc.), thereby making analysis of the behavior of the model easier. 
We gather the graph data that have 25 nodes and the features shown in Table~\ref{dataset}.
In the Table~\ref{dataset}, a conditional vector of a scaling exponent $F_\mathrm{deg}(G_i)$ of a degree distribution and a clustering coefficient $F_\mathrm{cluster}(G_i)$ as [$F_\mathrm{deg}(G_i)$\;$F_\mathrm{cluster}(G_i)$]. 

\vspace{-3ex}
\begin{table}[H]
	\begin{center}
		\caption{Training dataset}
		\label{dataset}
    \vspace{-3ex}
		\begin{tabular}{c|c|c|c}
			conditional vector&scaling exponent&clustering coefficient&\# of data\\\hline
      $[-1.1\;\, 0.2]$&-1.1&0.2&400\\
      $[-0.8\;\, 0.4]$&-0.8&0.4&400\\
      $[-0.5\;\, 0.6]$&-0.5&0.6&400
		\end{tabular}
	\end{center}
\end{table}
\vspace{-3.5ex}

The parameters in the model are set as follows. 
The dimension of the hidden layer of LSTM, the embedded layer, and the linear layer to convert the output to the parameters of the normal distribution are set to 256, 128, and 20, respectively. 
The initial learning rate, the weight decay, and the threshold of gradient clipping are set to 0.001, 0.01, and 0.015, respectively.
The number of mini-batch is 60 and the number of epochs is 400.
\begin{figure}[tb]
	\vspace{-4ex}
	\centering
	\includegraphics[width=0.75\linewidth]{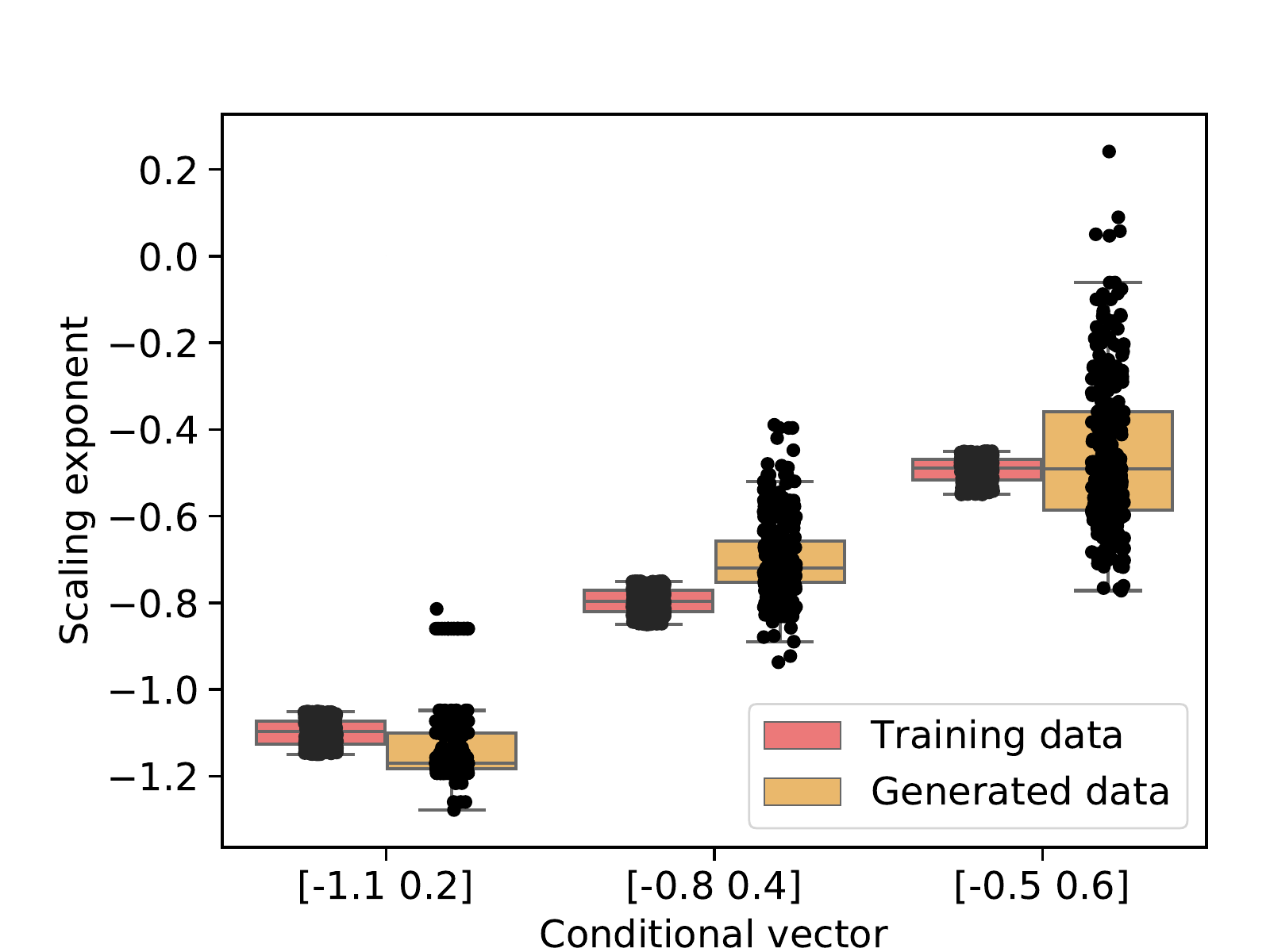}
	\vspace{-2ex}
	\caption{Scaling exponent of degree distributions}
	\label{eval1_degree}
	\vspace{-3ex}
\end{figure}

Fig.~\ref{eval1_degree} shows the distribution of the scaling exponents of the training data and the generated data.
The vertical axis represents the value of the scaling exponents. 
We can confirm that the scaling exponents are distributed around each value (i.e., -1.1, -0.8, and -0.5) specified by the conditional vectors.

\section{Conclusions}\label{conclusion}

We proposed a tunable model that can generate a graph with features specified by a conditional vector, using CVAE. 
It was confirmed that the proposed model can generate a graph with specified features through the experiments. 
We have a plan to verify the applicability of our model in a real dataset. 
The verification includes additional experiments for a wide variety of multi-aspect features, graph scale, and so on.

\bibliographystyle{IEEEtran}\bibliography{list}

\end{document}